\documentclass{article} % For LaTeX2e
\usepackage{iclr2027_conference,times}

\usepackage{mathtools}
\usepackage{listings}
\usepackage{multirow}
\usepackage{wrapfig}
\usepackage{url}
\usepackage{xcolor}
\usepackage{enumitem}
\usepackage{capt-of}
\usepackage{booktabs}
\usepackage{subcaption}
\usepackage{calrsfs}
\usepackage{tikz}

\newcommand{\system}{\textsc{\mbox{ACID}}\xspace}

\usepackage{xspace}
\usepackage{soul}

\lstdefinelanguage{JSON}{
    basicstyle=\ttfamily,
    showstringspaces=false,
    breaklines=true,
    string=[s]{"}{"},
    stringstyle=\color{blue},
    numbers=left,
    numberstyle=\tiny,
    backgroundcolor=\color{gray!10}
}

\lstdefinelanguage{JavaScript}{
  keywords={typeof, new, true, false, catch, function, return, null, switch, var, if, in, while, do, else, case, break, let, const},
  keywordstyle=\color{blue}\bfseries,
  ndkeywords={class, export, boolean, throw, implements, import, this},
  ndkeywordstyle=\color{teal}\bfseries,
  identifierstyle=\color{black},
  sensitive=false,
  comment=[l]{//},
  morecomment=[s]{/*}{*/},
  commentstyle=\color{gray}\ttfamily,
  stringstyle=\color{red}\ttfamily,
  morestring=[b]',
  morestring=[b]"
}

\lstdefinestyle{customc2}{
  language=JavaScript,
  basicstyle=\ttfamily\small,
  keywordstyle=\color{blue},
  stringstyle=\color{red},
  commentstyle=\color{gray},
  numbers=left,
  numberstyle=\tiny,
  breaklines=false,
  showstringspaces=false
}

\usepackage{graphicx}

\usepackage[utf8]{inputenc} % allow utf-8 input
\usepackage[T1]{fontenc}    % use 8-bit T1 fonts
\usepackage{hyperref}       % hyperlinks
\definecolor{iclrlink}{RGB}{0,84,166}
\hypersetup{colorlinks=true, linkcolor=iclrlink, citecolor=iclrlink,
            urlcolor=iclrlink, filecolor=iclrlink}
\usepackage{url}            % simple URL typesetting
\usepackage{booktabs}       % professional-quality tables
\usepackage{amsfonts}       % blackboard math symbols
\usepackage{nicefrac}       % compact symbols for 1/2, etc.
\usepackage{microtype}      % microtypography
\usepackage{xcolor}         % colors
\usepackage{amsmath}
\usepackage{multirow}
\usepackage{algorithm}
\usepackage{algpseudocode}
\usepackage{subcaption}

\title{One Threshold Is Not Enough: \\ Prompt-Invariant Caching Schedules \\ for Video Diffusion}

\author{%
  Om Agrawal \\
  UT Austin \\
  \And
  Saurabh Agarwal \\
  UT Austin \\
  \And
  Aditya Akella \\
  UT Austin \\
}

\iclrfinalcopy % arXiv preprint: de-anonymised, no line numbers
\begin{document}

\maketitle
\lhead{Preprint}

\begin{abstract}
Video diffusion models remain expensive because denoising repeatedly evaluates a costly model backbone. Dynamic caching methods like TeaCache, EasyCache, and DiCache reduce this cost by reusing prior outputs while an accumulated drift signal remains below a threshold $\tau$. Although existing methods improve the drift signal, they hold $\tau$ fixed throughout denoising. We show that fixed thresholds are structurally suboptimal: varying $\tau$ across denoising achieves quality--latency tradeoffs unattainable by any fixed threshold across three caching methods with distinct drift signals. We further show that, surprisingly, each method's drift trajectory is nearly prompt-invariant across seven method--model combinations. The shape of the drift trajectory is determined by the model and caching method; therefore, a threshold schedule can be calibrated offline and reused across generations. Based on these findings, we present \system{} (Adaptive Caching for vIDeo generation), which uses a low threshold during critical regions where drift changes rapidly and a high threshold across stable regions. \system{} identifies these regions offline from the drift signal's second derivative, requires no training, and adds negligible runtime overhead. Across TeaCache, EasyCache, and DiCache on HunyuanVideo, Wan~2.1, and CogVideoX, \system{} pushes the speed--quality Pareto frontier beyond fixed thresholds. On TeaCache with HunyuanVideo, it achieves $2.16\times$ speedup over no caching and $38\%$ additional speedup over a conservative fixed threshold, with less than $0.3$ dB PSNR, $0.01$ SSIM, and $0.01$ LPIPS degradation relative to that fixed-threshold configuration.
\end{abstract}

\section{Introduction}

Video generation models~\citep{zheng2024opensora, ma2024latte, brooks2024sora, wang2025wan, hong2023cogvideo, kong2024hunyuanvideo} are rapidly being used to generate realistic videos~\citep{zheng2025vbench2}; however, their slow inference speed and high compute requirements (3 to 23 minutes for ${\sim}$5-second videos on a single A100 GPU) pose an impediment to their wide adoption~\citep{li2024snapfusion}.
As these models scale to generate even higher resolutions, the challenges associated with slow inference speed and high computation requirements will be further exacerbated.
The primary reason for these slow inference speeds and high computation needs is the
core inference loop of these video generation models~\citep{zheng2024opensora, ma2024latte, brooks2024sora, wang2025wan, hong2023cogvideo, kong2024hunyuanvideo}: a computationally expensive diffusion backbone that runs a sequential denoising procedure, dominating generation time.

Training-free methods accelerate video generation by reducing the cost of denoising~\citep{zhang2025spargeattn, xi2025sparsevideogen, ren2025grouping, lu2022dpmsolver, liu2025teacache}. Caching methods are appealing because they are lightweight and easy to deploy. These methods exploit similarity across diffusion steps to reuse previously computed representations. Early caching methods reuse representations at fixed intervals. Recent methods instead adapt their caching decisions at runtime~\citep{kahatapitiya2024adacache, liu2025teacache, zhou2025easycache, liu2025taylorseers, bu2025dicache}. 
At each step, they estimate the change in model state and use it to decide whether to reuse a cached representation or recompute it.
% To alleviate the computational bottleneck of video generation, several training-free acceleration approaches have been proposed~\citep{zhang2025spargeattn, xi2025sparsevideogen, ren2025grouping, lu2022dpmsolver, liu2025teacache}. Among these approaches, caching-based methods are appealing because of their lightweight computational requirements and ease of deployment. These methods exploit the similarity between intermediate states across diffusion steps and reuse previously computed representations to reduce redundant computation. Caching strategies have evolved from early uniform schemes that reuse intermediate states at fixed intervals to more sophisticated adaptive methods~\citep{kahatapitiya2024adacache, liu2025teacache, zhou2025easycache, liu2025taylorseers, bu2025dicache}. These adaptive methods determine when and what to cache based on the estimated similarity at each step, allowing the caching policy to account for the dynamic behavior of the diffusion process.

Popular caching methods, including TeaCache~\citep{liu2025teacache}, EasyCache~\citep{zhou2025easycache}, and DiCache~\citep{bu2025dicache}, share the same basic structure. Each accumulates an estimate of the change in model output. When the accumulated drift remains below a user-defined threshold $\tau$, the method skips the expensive model evaluation and reuses the previous output with a correction. When the drift exceeds $\tau$, the method runs the full model and resets the accumulator.

Within this shared structure, successive methods differ primarily in how they estimate drift. TeaCache uses changes in the model input. EasyCache instead estimates the transformation rate from input change to output change. DiCache uses shallow features from the model's early layers. Despite these different drift signals, all three compare the accumulated drift against a single user-defined threshold $\tau$. This threshold remains fixed throughout denoising.

% Popular caching methods like TeaCache~\citep{liu2025teacache}, EasyCache~\citep{zhou2025easycache}, and DiCache~\citep{bu2025dicache}, all share a common structure: they use an accumulated drift metric that estimates the change in model output. If this accumulated drift is below a user-provided threshold $\tau$, they skip the expensive model run and reuse the previous output with some correction. If the accumulated drift exceeds the threshold, they run the full model and reset the accumulated drift to zero.

\begin{figure}[t]
    \centering
    \includegraphics[width=1\linewidth]{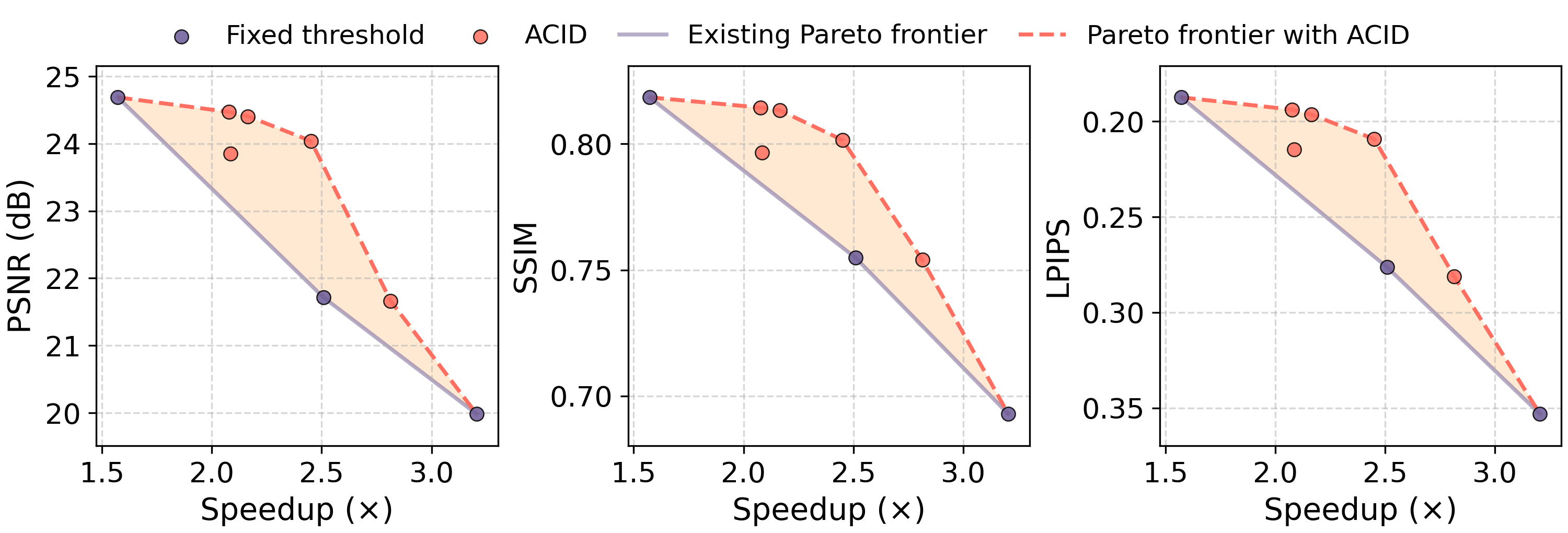}
    \caption{TeaCache on HunyuanVideo. \system reaches quality--latency operating points beyond those attainable by any fixed-threshold configuration.}
    \label{fig:pareto-good}
\end{figure}

% Within the shared structure of caching methods , successive methods have competed almost entirely on the drift metric. TeaCache estimates drift from changes in the model's inputs, EasyCache replaces this with a runtime transformation rate relating output change to input change, and DiCache abandons input-level proxies altogether in favor of shallow feature probes taken from the model's own early layers. Each is a different estimate of the same quantity. Yet across all three, the threshold that these estimates are compared against is a single number, supplied by the user, and held constant for every step of the denoising process.
Choosing $\tau$ exposes a tradeoff between output quality and inference latency. A high threshold caches aggressively, reducing latency at the cost of output quality. A low threshold preserves quality but provides less acceleration. This raises a natural question: \textit{Do fixed thresholds achieve the best possible quality--latency tradeoff?} We show that they do not. Varying $\tau$ across denoising reaches operating points unattainable by any fixed threshold, as Figure~\ref{fig:pareto-good} shows. This result holds for TeaCache, EasyCache, and DiCache despite their different drift signals. The limitation therefore lies not only in how drift is estimated, but also in applying a single threshold throughout denoising. Threshold scheduling provides a complementary axis for improving the quality--latency tradeoff.

Varying $\tau$ requires deciding when to change it. At first glance, this appears to require a runtime policy that adapts the threshold to each prompt. However, across 98 prompts and seven method--model combinations, each method's drift signal follows a nearly identical trajectory across prompts (Figures~\ref{fig:prompt-invariance} and~\ref{fig:prompt-invariance-appendix}). Because the trajectory depends on the model and caching method, both fixed at deployment, the threshold schedule can be computed once offline and reused across generations.

% Varying $\tau$ requires deciding when to change it. On first thought, we might want to come up with  a runtime policy could make this decision for each prompt, but we surprisingly observed that  that it is unnecessary. Across 98 prompts and seven method--model combinations, each method's drift signal follows a nearly identical trajectory across prompts (Figures~\ref{fig:prompt-invariance} and~\ref{fig:prompt-invariance-appendix}). Its shape instead depends on the model and caching method, both fixed at deployment. The threshold schedule can therefore be computed once offline and reused for every generation.

% Exploiting a second axis normally costs something, typically runtime machinery to decide how to move along it. Here it costs almost nothing. We measure each method's drift signal across 98 prompts on all seven method and model combinations we study (Figures~\ref{fig:prompt-invariance} and~\ref{fig:prompt-invariance-appendix}), and find that the signal traces essentially the same trajectory regardless of what is being generated. Its shape instead reflects the model and the caching method, both of which are fixed at deployment time. The schedule for varying $\tau$ can therefore be computed once, offline, and reused for every subsequent generation, rather than being inferred at runtime for each prompt
We attribute the gains from threshold scheduling to \emph{critical steps}, regions of denoising where the drift signal changes rapidly. Aggressive caching during these regions causes disproportionate quality loss. We therefore use a low threshold at critical steps and a high threshold elsewhere. Because critical steps occur in consistent regions of the trajectory, their boundaries can be identified offline and reused across generations.

% We attribute the power of multi-threshold caching schemes to the existence of \emph{critical steps} in the sequential denoising procedure, timesteps at which the drift signal changes rapidly and the output is most sensitive to approximation. By avoiding aggressive caching during these critical steps, our method mitigates the quality loss that fixed-threshold caching methods typically exhibit. Because these steps occupy consistent regions of the trajectory, one can design a caching scheme that accounts for them ahead of time and sets the threshold accordingly.

We demonstrate these findings with \system{} (Adaptive Caching for vIDeo generation), a minimal threshold schedule. \system{} applies a low threshold $\tau_{\text{low}}$ at critical steps and a high threshold $\tau_{\text{high}}$ across the stable interior. It identifies these regions offline using the second derivative of the caching method's drift signal. This simple design is training-free, signal-agnostic, and adds negligible runtime overhead.

Across three caching methods and three video diffusion models, \system{} pushes the speed--quality Pareto frontier beyond fixed thresholds. On TeaCache with HunyuanVideo, it achieves a $2.16\times$ speedup over no caching and $38\%$ additional speedup over a conservative fixed threshold, with less than $0.3$ dB PSNR, $0.01$ SSIM, and $0.01$ LPIPS degradation relative to the latter.

% We demonstrate both findings with \system (Adaptive Caching for vIDeo generation), our minimal instantiation of a varying threshold. \system applies a low threshold $\tau_{\text{low}}$ at critical steps and a high threshold $\tau_{\text{high}}$ across the stable interior of the trajectory, with the boundaries between these regions located offline from the second derivative of the caching method's own drift signal. \system is training-free, signal-agnostic, and adds negligible runtime overhead. We do not claim this is the optimal way to vary $\tau$; it is deliberately the simplest one, chosen to show that the gains come from varying the threshold at all rather than from any particular machinery.
% 
% Our experiments show that \system pushes the speed--quality Pareto frontier beyond any fixed threshold, achieving reduced inference latency while maintaining output quality on video generation tasks. In particular, on TeaCache and HunyuanVideo, \system achieves up to $2.16\times$ speedup over the no-caching baseline, and up to $38\%$ additional speedup over the conservative fixed-threshold baseline with negligible ($<0.3$ dB PSNR, $<0.01$ SSIM, $<0.01$ LPIPS) quality degradation.

We summarize our contributions below:
\begin{itemize}
    \item We show that the caching threshold $\tau$, held fixed by every dynamic caching method for video diffusion, is an independent axis of improvement from the drift signal those methods have competed on. Varying $\tau$ across denoising reaches quality--latency operating points that no fixed $\tau$ attains, and it does so for TeaCache, EasyCache, and DiCache alike, despite their unrelated drift signals. This indicates the limitation is structural rather than a deficiency of any single signal.

    \item We show, across 98 prompts and all seven method--model combinations we evaluate, that each method's drift signal traces essentially the same trajectory regardless of the prompt, and that its shape is instead characteristic of the model and the caching method. Since both are fixed at deployment, the threshold schedule can be computed once offline rather than inferred at runtime.

    \item We identify \emph{critical steps} in the sequential denoising procedure, timesteps at which the drift signal changes rapidly and aggressive caching disproportionately degrades quality, and show that they occupy consistent regions of the trajectory that can be located offline from the second derivative of the drift signal.

    \item We demonstrate both findings with \system, a minimal threshold schedule requiring no training, no changes to the underlying caching method, and negligible runtime overhead. Across three caching methods and three video diffusion models, \system pushes the achievable speed--quality Pareto frontier beyond any fixed threshold, with the largest gains on TeaCache and HunyuanVideo.
\end{itemize}

\section{Related Work}

\paragraph{Diffusion Models.}
While U-Net-based diffusion models like Stable Diffusion~\citep{rombach2022ldm} and SDXL~\citep{podell2023sdxl} successfully catalyzed high-quality image and video generation~\citep{guo2023animatediff, chen2024videocrafter2, blattmann2023svd}, their limited architectural capacity bottlenecks the training and deployment of massive models. To eliminate this constraint, recent frameworks have rapidly shifted toward the Diffusion Transformer (DiT) backbone, exploiting its structural flexibility to support next-generation foundation models~\citep{esser2024scaling, blackforestlabs2024flux, chen2023pixart, peebles2023scalable}. However, despite their theoretical promise, scaling these DiT backbones introduces substantial computational overhead.

\paragraph{Diffusion Model Acceleration.}
As DiT backbones scale in capacity and complexity, their runtime increases proportionately, exacerbating the notoriously slow inference speeds of diffusion models and impeding real-world deployment. To alleviate this latency bottleneck, several acceleration paradigms have emerged. Recent work has extensively explored efficient attention~\citep{dao2022flashattention}, sparse attention~\citep{xi2025sparsevideogen, ren2025grouping}, distillation~\citep{meng2023distillation, sauer2024adversarial, wang2023videolcm}, quantization~\citep{li2024svdquant, chen2025qdit}, and improved SDE/ODE solvers~\citep{lu2022dpmsolver}. Alongside these methods, caching-based feature reuse strategies offer a particularly compelling, orthogonal path forward, which we detail next.

\paragraph{Caching-based feature reuse strategies.}
Caching methods have gained attention recently due to their lightweight nature. DeepCache~\citep{ma2023deepcache} and Faster Diffusion~\citep{li2023fasterdiffusion} reuse U-Net features across timesteps to reduce computational redundancy. FORA~\citep{selvaraju2024fora}, $\Delta$-DiT~\citep{chen2024deltadit}, and Learning-to-Cache~\citep{ma2024learningtocache} extend this idea to transformer-based architectures. PAB~\citep{zhao2024pab} broadcasts attention features to later timesteps in a pyramid-style method based on various block characteristics. AdaCache~\citep{kahatapitiya2024adacache} modifies residual reuse methods based on content complexity. FasterCache~\citep{lv2024fastercache} proposes to cache for both the conditional and unconditional branch of Classifier Free Guidance~\citep{ho2022cfg}. TeaCache~\citep{liu2025teacache}, EasyCache~\citep{zhou2025easycache}, and DiCache~\citep{bu2025dicache} represent the most recent state-of-the-art dynamic caching approaches; we examine their shared structure and its key limitation in the paragraphs below.

\paragraph{The Fixed Threshold of Dynamic Feature Caching Methods.}
Most recent dynamic caching-based feature reuse strategies make caching decisions based on the diffusion process's state at runtime. Methods like TeaCache, EasyCache, and DiCache share a common algorithmic skeleton: a running accumulated drift variable is initialized to zero, an estimate of output change is added at each step, and this accumulated value is compared against a fixed threshold $\tau$. When the accumulated drift is below $\tau$, the expensive model forward pass is skipped and the previous output is reused with some correction. When the accumulated drift exceeds $\tau$, the full model is run and the accumulator is reset. The methods differ in how they estimate drift and how they reconstruct the output on skipped steps, which we delve into below.

\paragraph{Warmup Phases.}
Each of these methods runs the full model for a hardcoded number of initial steps before caching begins. This is effectively a coarse threshold schedule, with $\tau = 0$ early and $\tau > 0$ afterward, where the phase boundary is fixed per model and never varied per prompt. The practice is treated as an implementation detail rather than a design principle, and the threshold is held constant everywhere outside this prefix.

\paragraph{TeaCache.}
TeaCache estimates drift using the L1 distance between consecutive values of the Timestep Embedding Modulated Noisy Input (TEMNI), a combination of the text embedding, timestep embedding, and current noisy input. Rather than measuring how much the model output changed, which would require running the model, TeaCache measures how much the model's inputs changed, exploiting the empirical observation that large input changes correlate with large output changes. On skipped steps, TeaCache reuses the previous output with a stored residual correction.

\paragraph{EasyCache.}
EasyCache estimates drift using the model's transformation rate $k$, defined as the ratio of output change to input change at the most recently computed step. The authors observe that $k$ is unstable early in the denoising process but quickly stabilizes, reflecting the model settling into a roughly linear input-output regime. Under this regime, EasyCache estimates the current step's output change as $k \times \Delta\text{input} / \|v_{t-1}\|$, where $\Delta\text{input}$ is the current input change, $\|v_{t-1}\|$ is the magnitude of the previous output, and $k$ is the transformation rate of the last fully computed step. This normalized estimate is accumulated against $\tau$. On skipped steps, EasyCache reconstructs the output as the current input plus the offset $\Delta = \text{output} - \text{input}$ from the last fully computed step, which is justified because a stable $k$ implies the model's additive contribution is approximately constant.

\paragraph{DiCache.}
DiCache replaces the input-level proxy used by TeaCache and EasyCache with a feature-level proxy. At each step, DiCache runs only the first few transformer blocks of the model as a shallow probe and measures how much those features changed. Because these early layers are cheap to compute, this provides a direct estimate of whether the full model's output would change significantly, without paying the cost of a full forward pass. On skipped steps, DiCache uses the trajectory of shallow probe features across multiple recent steps to interpolate a better estimate of the current deep output.

Despite these differences, all three methods share a critical limitation: the fixed user-provided threshold is held constant throughout the entire denoising process. This means the caching schedule cannot respond to the diffusion process's varying dynamics. Note that all three improve the drift estimate while leaving the threshold untouched; our work shows that the threshold is itself an independent axis of improvement, and that because the drift signal is consistent across prompts, a schedule for varying it can be computed once offline rather than inferred at runtime.

\section{\system}
In this section, we introduce \system, a method for varying caching frequency across the denoising process.

\subsection{Analysis}
\label{sec:analysis}

\begin{figure}[t]
    \centering
    \includegraphics[width=\linewidth]{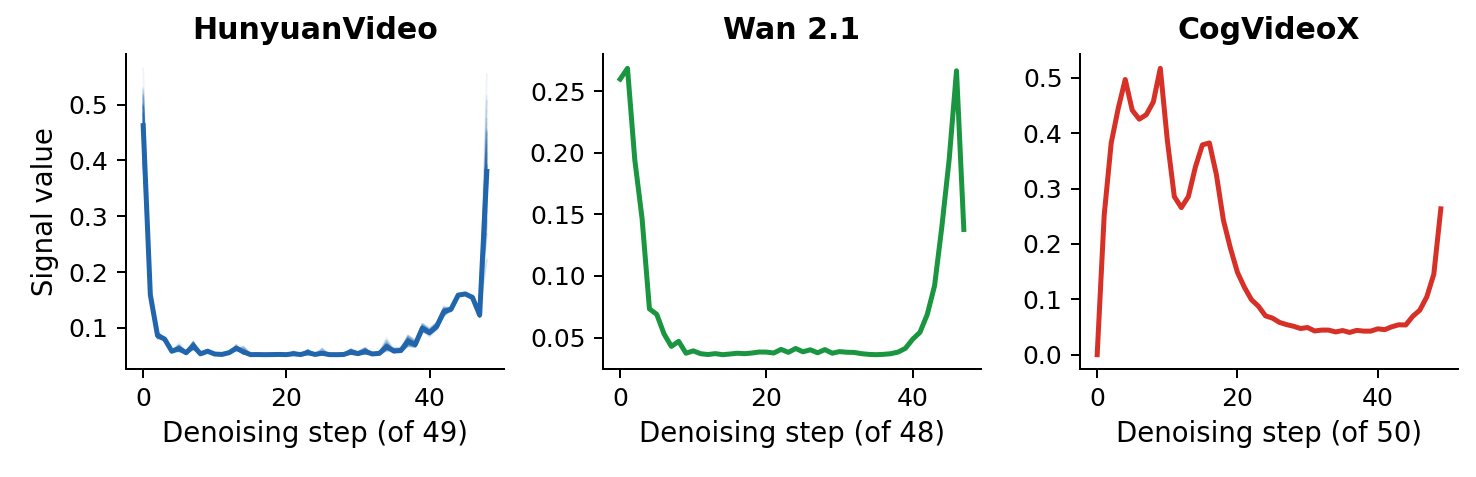}
    \caption{TeaCache drift signal across 98 T2V-CompBench prompts; faint lines are individual prompts, the bold line is the mean. Trajectories are nearly indistinguishable within a panel but differ across panels, showing the signal reflects the model rather than the prompt. EasyCache and DiCache appear in Figure~\ref{fig:prompt-invariance-appendix}.}
    \label{fig:prompt-invariance}
\end{figure}

\paragraph{Two properties of caching signals.}
To understand the limitations of fixed-threshold caching, we analyze how caching signals evolve throughout the denoising process. Figures~\ref{fig:prompt-invariance} and~\ref{fig:prompt-invariance-appendix} show each method's drift signal measured across 98 T2V-CompBench prompts, for every method--model combination we evaluate. Two properties are apparent.

First, the signal is consistent across prompts. The 98 prompts are drawn evenly from the seven compositional categories of T2V-CompBench~\citep{sun2025t2vcompbench}. Within any single panel, the trajectory produced by a visually complex, fast-moving scene is nearly indistinguishable from that of a static, minimal one. 

Second, the shape of the signal is characteristic of the model and the caching method. HunyuanVideo and Wan~2.1 under TeaCache exhibit a U-shaped trajectory, falling steeply from a high initial value, remaining flat through the middle of denoising, and rising again near the end. CogVideoX under TeaCache shows a visibly different structure, with an extended high-signal region early in denoising. The same model produces different shapes under different caching methods, since each method measures a different quantity.

Taken together, these properties mean the trajectory of the drift signal is fixed once the model and the caching method are chosen, both of which are decided at deployment time rather than per generation. Anything we derive from the shape of this trajectory can therefore be computed ahead of time.

We note that the drift signal is a proxy: each method measures something cheap in place of the model output change it cannot observe without running the model. Figure~\ref{fig:compare_prompts} in Appendix~\ref{sec:appendix-signal} compares TeaCache's $\Delta$TEMNI against the actual relative change in model output on HunyuanVideo, confirming that the proxy tracks the quantity it is meant to estimate.

\paragraph{Why one threshold is not enough.}
As Figure~\ref{fig:prompt-invariance} shows, there are certain parts of the diffusion process where the drift signal is relatively high, and other parts where it is low. Dynamic feature caching methods already take this into account: when the predicted $\Delta$output is greater, they cache less often, because they hit the fixed threshold more often, and when the predicted $\Delta$output is smaller, they cache more often. But while dynamic caching methods do find a better speed--quality tradeoff than the no-caching baseline, we find that this tradeoff is not optimal: the degree to which they cache more in stable zones or cache less in unstable zones is suboptimal, a limitation imposed by the single fixed threshold. We find that using different thresholds for different parts of the diffusion process, such as low thresholds for unstable regions and high thresholds for stable regions, leads to a better speed--quality tradeoff.

We could attribute the gains from varying the threshold to a suboptimal metric for TeaCache, but we importantly find the same gains for EasyCache and DiCache as well, which use entirely different caching signals. This suggests the limitation is structural: any method that applies a single fixed threshold throughout the denoising process will face this suboptimality, regardless of the signal it uses.

\paragraph{Locating the transition points.}
The previous observations establish that multiple thresholds help, but not where along the trajectory the threshold should change. We address this in two stages: first establishing that some transition points outperform every fixed threshold, then asking what property of the signal predicts where those points fall.

For the first stage, we ran a calibration study using TeaCache on HunyuanVideo, Wan~2.1, and CogVideoX. For each model we evaluated a low fixed threshold, a high fixed threshold, and a range of multi-threshold configurations in which we varied the steps at which the threshold changes. Figure~\ref{fig:metric-fd} in Appendix~\ref{sec:appendix-signal} shows the $\Delta$TEMNI signal for each model, with dashed lines marking the transition points at which a multi-threshold configuration achieved a better speed--quality tradeoff than both fixed-threshold baselines. Such points exist for all three models, confirming that the gains are not specific to a particular model's signal shape.

For the second stage, we asked what distinguishes these transition points from any other step. Examining the second derivative of the signal (Figure~\ref{fig:metric-sd}), we found that they coincide with the steps at which the second derivative stabilizes, marking where the signal passes from a rapidly changing regime into a stable one. This gives a criterion for placing the transitions that depends only on the shape of the drift signal, which the previous properties established can be measured offline.

This criterion was derived from TeaCache alone. We applied it unchanged to EasyCache and DiCache, whose drift signals measure entirely different quantities, and obtained schedules that expand the quality--latency frontier in every method--model combination we evaluate (Appendix~\ref{sec:appendix-pareto}). That a criterion calibrated on one signal transfers to two unrelated ones supports the view that it is capturing a property of the denoising process rather than an artifact of any single metric.

\subsection{Threshold Schedules}

\system extends the single-threshold framework of methods like TeaCache, EasyCache, and DiCache with a threshold schedule. Rather than applying a single fixed threshold $\tau$ at every step, we partition the denoising trajectory into regions and assign a threshold to each. We use two threshold values: a low threshold $\tau_{\text{low}}$, which caches conservatively and preserves fidelity, and a high threshold $\tau_{\text{high}}$, which caches aggressively and reduces latency.

The regions follow the structure of the drift signal. Steps at which the signal changes rapidly are \emph{critical steps}, where approximating the model output is most damaging to quality, and these receive $\tau_{\text{low}}$. Steps in the stable interior of the trajectory, where the signal is flat and successive outputs differ little, receive $\tau_{\text{high}}$. For the U-shaped signals of HunyuanVideo and Wan~2.1 this yields a low-high-low schedule, since the signal changes rapidly at both ends of denoising. For signals with a different structure the schedule follows accordingly.

Because the drift signal is fixed once the model and caching method are chosen, the boundaries between these regions are determined once, offline, by the calibration procedure of Algorithm~\ref{alg:calibration}, and reused for every subsequent generation. At inference time (Algorithm~\ref{alg:schedule_caching}) the underlying method's accumulate-and-compare loop is unchanged except that the threshold it compares against is read from the schedule rather than fixed. \system therefore requires no retraining, no modification to how a method estimates drift or reconstructs skipped outputs, and adds negligible runtime overhead.

\begin{algorithm}[t]
\caption{Offline Schedule Calibration}
\label{alg:calibration}
\begin{algorithmic}[1]
\Require Model $f_\theta$, base method $\mathcal{M}$, calibration prompts $P$, sampling steps $T$
\For{each prompt $p \in P$}
    \State Run $f_\theta$ without caching, recording $s_t \leftarrow \mathcal{M}.\text{signal}(x_t, t, c_p)$ for $t = T, \ldots, 0$
\EndFor
\State $\bar{s} \leftarrow$ mean signal trajectory over $P$ \hfill \textit{// prompt-invariant by Figure~\ref{fig:prompt-invariance}}
\State $\bar{s}'' \leftarrow$ second difference of $\bar{s}$
\State $R_{\text{crit}} \leftarrow \{t : \bar{s}'' \text{ has not yet stabilized}\}$ \hfill \textit{// critical steps at trajectory ends}
\State \Return $R_{\text{crit}}$
\end{algorithmic}
\end{algorithm}

\begin{algorithm}[t]
\caption{Inference with a Threshold Schedule}
\label{alg:schedule_caching}
\begin{algorithmic}[1]
\Require Sampling steps $T$, base method $\mathcal{M}$, thresholds $\tau_{\text{low}}$, $\tau_{\text{high}}$, critical step set $R_{\text{crit}}$
\State Initialize accumulated drift $\Sigma \leftarrow 0$
\For{$t = T$ \textbf{to} $0$}
    \If{$\mathcal{M}.\text{warmup}(t)$}
        \State $y_t \leftarrow f_\theta(x_t, t, c)$; \; $\mathcal{M}.\text{update}(y_t)$ \hfill \textit{// run full model during warmup}
    \Else
        \State $s_t \leftarrow \mathcal{M}.\text{signal}(x_t, t, c)$ \hfill \textit{// compute drift signal via $\mathcal{M}$}
        \State $\Sigma \leftarrow \Sigma + s_t$
        \If{$t \in R_{\text{crit}}$}
            \State $\tau \leftarrow \tau_{\text{low}}$ \hfill \textit{// critical step: cache conservatively}
        \Else
            \State $\tau \leftarrow \tau_{\text{high}}$ \hfill \textit{// stable interior: cache aggressively}
        \EndIf
        \If{$\Sigma \leq \tau$}
            \State $y_t \leftarrow \mathcal{M}.\text{reuse}(x_t, t)$ \hfill \textit{// skip full model; reconstruct via $\mathcal{M}$}
        \Else
            \State $y_t \leftarrow f_\theta(x_t, t, c)$; \; $\mathcal{M}.\text{update}(y_t)$; \; $\Sigma \leftarrow 0$ \hfill \textit{// run full model, reset}
        \EndIf
    \EndIf
\EndFor
\end{algorithmic}
\end{algorithm}

\section{Experiments}

\subsection{Experimental Setup}

\paragraph{Base models.}
We evaluate on three open-source video generation diffusion models: HunyuanVideo T2V 720P~\citep{kong2024hunyuanvideo}, Wan~2.1 T2V 1.3B~\citep{wang2025wan}, and CogVideoX 1.5-5B~\citep{yang2024cogvideox}.

\paragraph{Compared methods.}
We apply our threshold schedule framework on top of three base caching methods: TeaCache (evaluated on HunyuanVideo, Wan~2.1, CogVideoX), EasyCache (evaluated on HunyuanVideo, Wan~2.1), and DiCache (evaluated on HunyuanVideo, Wan~2.1). EasyCache and DiCache are not evaluated on CogVideoX because neither method provides an official implementation for that model; we restrict our evaluation to officially supported configurations to avoid confounding results from self-implemented ports. For each method--model combination, we compare a no-caching baseline, the fixed-threshold method with several threshold values spanning a low-to-high range, and our scheduled-threshold variant \system with several low/high threshold values.

\paragraph{Evaluation metrics.}
Following TeaCache, we evaluate along two dimensions: inference efficiency and visual quality. For inference efficiency, we report inference latency and speedup, defined as the ratio of the no-caching baseline latency to the method latency for the same model. For visual quality, we use LPIPS (perceptual similarity), PSNR (peak signal-to-noise ratio), and SSIM (structural similarity), computed by comparing cached-method outputs against the no-caching baseline.

\paragraph{Implementation details.}
All experiments were carried out on NVIDIA A100 80GB PCIe GPUs with PyTorch. We selected 33 prompts from VBench's~\citep{huang2024vbench} prompt set: 3 prompts per group across 11 groups, where each group corresponds to one or more VBench evaluation dimensions, together spanning all 16 VBench dimensions. For each method--model combination, we generated 33 videos per configuration (one per prompt) across all evaluated threshold settings, using a fixed seed throughout. Separately, for the drift signal analysis,  we recorded signals over 98 T2V-CompBench~\citep{sun2025t2vcompbench} prompts for each of the seven method--model combinations, giving 686 signal-recording runs.

\subsection{Main Results}

\begin{table*}[t]
\centering
\footnotesize
\caption{Quantitative evaluation across caching methods and models. Per method--model, rows are: baseline (no caching), fixed low threshold, fixed high threshold, and our scheduled-threshold method (best Pareto-optimal configuration); the fixed thresholds are the low/high values used by the scheduled mode. Fidelity metrics are computed against the uncached baseline, and ``---'' marks the self-referential baseline. $\uparrow$/$\downarrow$: higher/lower is better.}
\label{tab:main_results}
\renewcommand{\arraystretch}{1.0}
\setlength{\tabcolsep}{4pt}
\begin{tabular}{lllccccc}
\toprule
\textbf{Method} & \textbf{Model} & \textbf{Mode} & \textbf{Latency} & \textbf{Speedup} & \textbf{PSNR~$\uparrow$} & \textbf{SSIM~$\uparrow$} & \textbf{LPIPS~$\downarrow$} \\
\midrule

\multirow{12}{*}{\textbf{TeaCache}}
  & \multirow{4}{*}{HunyuanVideo}
    & Baseline          & 1359s & 1.00$\times$ & ---   & ---   & ---   \\
  & & Fixed (lo, 0.1)   & 864s  & 1.57$\times$ & 24.69 & 0.819 & 0.187 \\
  & & Fixed (hi, 0.3)   & 424s  & 3.20$\times$ & 19.98 & 0.693 & 0.353 \\
  & & Scheduled         & 628s  & 2.16$\times$ & 24.41 & 0.813 & 0.196 \\
\cmidrule{2-8}
  & \multirow{4}{*}{Wan~2.1}
    & Baseline          & 212s  & 1.00$\times$ & ---   & ---   & ---   \\
  & & Fixed (lo, 0.05)  & 150s & 1.42$\times$ & 24.93 & 0.845 & 0.083 \\
  & & Fixed (hi, 0.2)   & 74s   & 2.86$\times$ & 17.10 & 0.538 & 0.335 \\
  & & Scheduled         & 108s  & 1.96$\times$ & 22.10 & 0.757 & 0.150 \\
\cmidrule{2-8}
  & \multirow{4}{*}{CogVideoX 1.5}
    & Baseline          & 1019s & 1.00$\times$ & ---   & ---   & ---   \\
  & & Fixed (lo, 0.1)   & 723s  & 1.41$\times$ & 37.32 & 0.958 & 0.029 \\
  & & Fixed (hi, 0.3)   & 465s  & 2.19$\times$ & 16.35 & 0.580 & 0.463 \\
  & & Scheduled         & 545s  & 1.87$\times$ & 32.02 & 0.930 & 0.058 \\
\midrule

\multirow{8}{*}{\textbf{EasyCache}}
  & \multirow{4}{*}{HunyuanVideo}
    & Baseline          & 1357s & 1.00$\times$ & ---   & ---   & ---   \\
  & & Fixed (lo, 0.025) & 659s  & 2.06$\times$ & 33.51 & 0.926 & 0.058 \\
  & & Fixed (hi, 0.045) & 531s  & 2.55$\times$ & 28.99 & 0.851 & 0.153 \\
  & & Scheduled         & 568s  & 2.39$\times$ & 30.62 & 0.894 & 0.094 \\
\cmidrule{2-8}
  & \multirow{4}{*}{Wan~2.1}
    & Baseline          & 216s  & 1.00$\times$ & ---   & ---   & ---   \\
  & & Fixed (lo, 0.020) & 125s  & 1.73$\times$ & 25.58 & 0.852 & 0.087 \\
  & & Fixed (hi, 0.040) & 102s  & 2.12$\times$ & 22.02 & 0.750 & 0.148 \\
  & & Scheduled         & 113s  & 1.92$\times$ & 24.90 & 0.833 & 0.100 \\
\midrule

\multirow{8}{*}{\textbf{DiCache}}
  & \multirow{4}{*}{HunyuanVideo}
    & Baseline          & 1265s & 1.00$\times$ & ---   & ---   & ---   \\
  & & Fixed (lo, 0.05)  & 659s  & 1.92$\times$ & 30.96 & 0.904 & 0.088 \\
  & & Fixed (hi, 0.10)  & 419s  & 3.02$\times$ & 20.15 & 0.720 & 0.328 \\
  & & Scheduled         & 620s  & 2.04$\times$ & 29.42 & 0.888 & 0.105 \\
\cmidrule{2-8}
  & \multirow{4}{*}{Wan~2.1}
    & Baseline          & 224s  & 1.00$\times$ & ---   & ---   & ---   \\
  & & Fixed (lo, 0.07)  & 99s   & 2.27$\times$ & 13.41 & 0.356 & 0.508 \\
  & & Fixed (hi, 0.225) & 56s   & 3.97$\times$ & 11.79 & 0.282 & 0.644 \\
  & & Scheduled         & 82s   & 2.73$\times$ & 13.29 & 0.355 & 0.541 \\
\bottomrule
\end{tabular}
\end{table*}

The Pareto plots in Appendix~\ref{sec:appendix-pareto} show the efficiency and visual quality tradeoff offered by fixed-threshold methods, and how our scheduled-threshold method expands the fixed-threshold Pareto frontier for each of the seven caching method and diffusion model combinations.

Table~\ref{tab:main_results} shows the efficiency and visual quality of notable scheduled mode points for each feature caching method and diffusion model combination, comparing them to the baseline mode, the fixed-threshold mode using the low threshold of the scheduled mode, and the fixed-threshold mode using the high threshold of the scheduled mode.

\paragraph{TeaCache.}
For TeaCache and HunyuanVideo, the scheduled mode achieves a 2.16$\times$ speedup, between the low-threshold fixed mode's speedup of 1.57$\times$ and the high-threshold fixed mode's speedup of 3.20$\times$, while offering fidelity scores very similar to those of the low-threshold mode.

For TeaCache and Wan~2.1, the scheduled mode achieves a speedup of 1.96$\times$, between the low-threshold fixed mode's speedup of 1.42$\times$ and the high-threshold fixed mode's speedup of 2.86$\times$, while retaining quality far closer to the low-threshold mode than to the high-threshold mode.

For TeaCache and CogVideoX, the scheduled mode achieves a speedup of 1.87$\times$, between the low-threshold fixed mode's speedup of 1.41$\times$ and the high-threshold fixed mode's speedup of 2.19$\times$, again retaining quality far closer to the low-threshold mode than to the high-threshold mode.

Figure~\ref{fig:pareto-teacache} shows how \system's scheduled modes surpass the Pareto frontier curves for all three diffusion models when used with TeaCache.

\paragraph{EasyCache.}
For EasyCache and HunyuanVideo, our scheduled mode achieves a speedup of 2.39$\times$, much closer to the high-threshold fixed mode's speedup of 2.55$\times$ than to the low-threshold fixed mode's speedup of 2.06$\times$, with quality metrics falling between those of the two fixed modes.

For EasyCache and Wan~2.1, our scheduled mode achieves a speedup of 1.92$\times$, falling between the low- and high-threshold fixed mode speedups of 1.73$\times$ and 2.12$\times$, while retaining quality far closer to the low-threshold mode than to the high-threshold mode.

Figure~\ref{fig:pareto-easycache} shows how several of \system's scheduled threshold modes surpass the Pareto frontier curves for both diffusion models when used with EasyCache.

We also note that EasyCache, by running the full model for the first few diffusion steps, already supports our main idea: varying the threshold across the diffusion process, in this case using a threshold of 0 (full computation) at the beginning and a higher threshold later, leads to a better quality--efficiency tradeoff.

\paragraph{DiCache.}
For DiCache and HunyuanVideo, the scheduled mode achieves a speedup of 2.04$\times$, falling between the low- and high-threshold speedups of 1.92$\times$ and 3.02$\times$, while achieving visual quality very close to the low-threshold mode and much better than the high-threshold mode.

For DiCache and Wan~2.1, the scheduled mode achieves a speedup of 2.73$\times$, falling between the low- and high-threshold speedups of 2.27$\times$ and 3.97$\times$, while achieving visual quality very close to the low-threshold mode and much better than the high-threshold mode.

Figure~\ref{fig:pareto-dicache} in Appendix~\ref{sec:appendix-pareto} shows how several of \system's scheduled modes surpass the Pareto frontier curve for both diffusion models when used with DiCache.

\section{Conclusion}

We showed two findings. First, the fixed caching threshold used by prior dynamic caching methods for video diffusion is a source of structural suboptimality. Second, the drift signal is a function of the model and the feature caching method rather than the prompt, so the schedule for varying the threshold can be computed once offline. We exploit both findings with \system, a minimal threshold schedule that caches conservatively at critical steps and aggressively across the stable interior. Experiments across three open-source video generation models and three caching methods confirm that \system achieves near-low-threshold visual quality with higher inference speedup.

\section*{AI Use Statement}

We used large language models to aid and polish the writing of this paper. All research contributions, including the experimental design, implementation, and analysis, are our own.

\appendix

\section{Drift Signal Analysis}
\label{sec:appendix-signal}

\begin{figure}[p]
    \centering
    % 2-panel PNGs are 5.55in wide vs 8.2in for the 3-panel one (same panel size),
    % so 0.677\linewidth keeps panels and text the same printed size as Figure~\ref{fig:prompt-invariance}.
    \begin{subfigure}[b]{0.677\linewidth}
        \includegraphics[width=\linewidth]{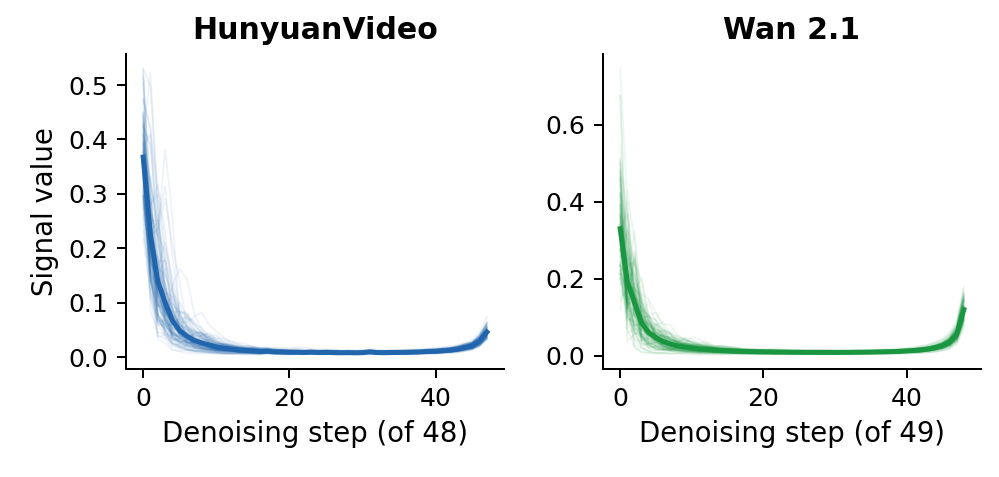}
        \caption{EasyCache}
        \label{fig:pi-easycache}
    \end{subfigure}\par
    \begin{subfigure}[b]{0.677\linewidth}
        \includegraphics[width=\linewidth]{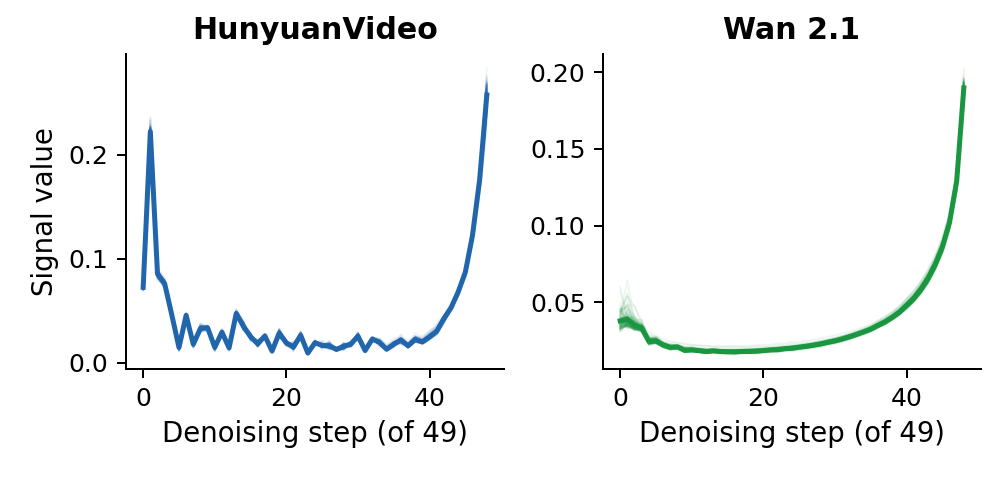}
        \caption{DiCache}
        \label{fig:pi-dicache}
    \end{subfigure}
    \caption{Drift signal for EasyCache and DiCache across 98 T2V-CompBench prompts, plotted as in Figure~\ref{fig:prompt-invariance}. The shape differs across methods, showing it is characteristic of the caching method as well as the model.}
    \label{fig:prompt-invariance-appendix}
\end{figure}

\begin{figure}[p]
    \centering
    \includegraphics[width=1\linewidth]{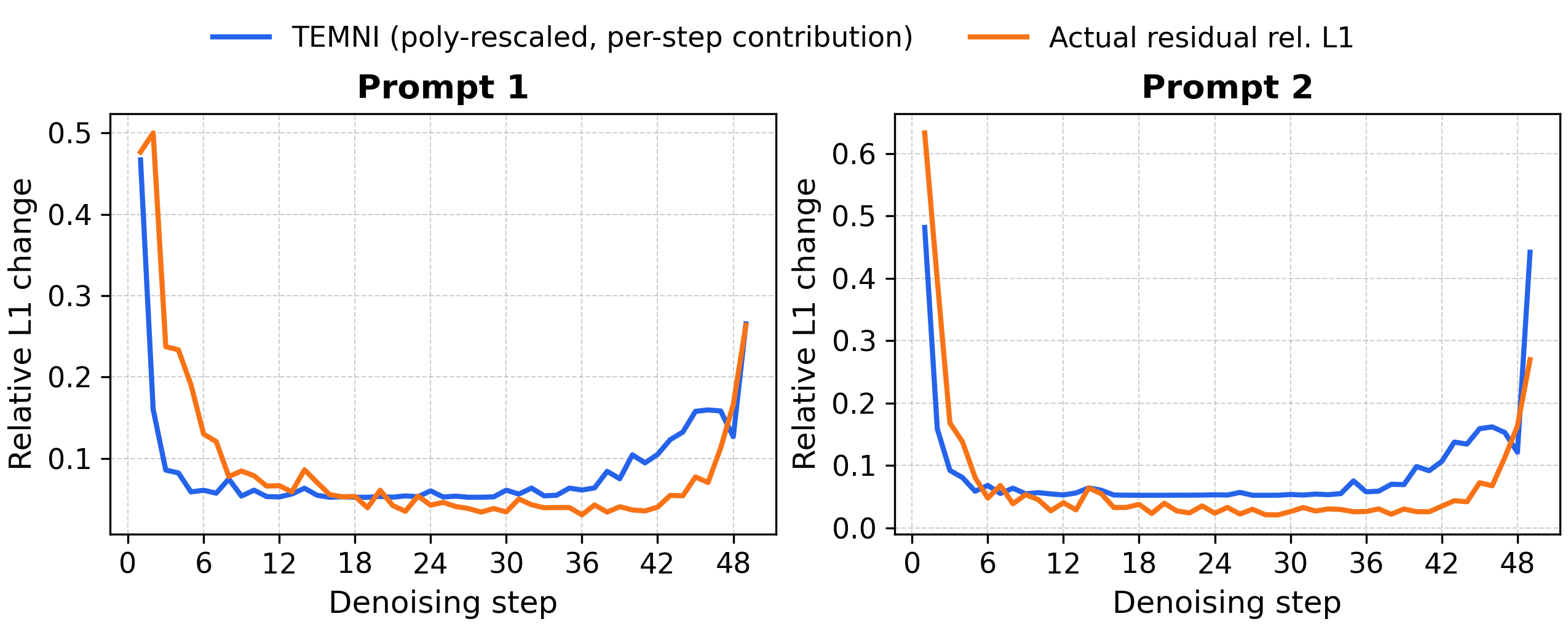}
    \caption{TeaCache's $\Delta$TEMNI proxy compared against the actual relative $L1$ change in model output on HunyuanVideo, for two prompts. \textbf{Prompt 1:} ``Two anthropomorphic cats in comfy boxing gear and bright gloves fight intensely on a spotlighted stage.'' \textbf{Prompt 2:} ``Static black and white shot of a single bare tree against an overcast sky, no movement, high contrast, minimalist composition.''}
    \label{fig:compare_prompts}
\end{figure}

\begin{figure}[p]
    \centering
    \includegraphics[width=1\linewidth]{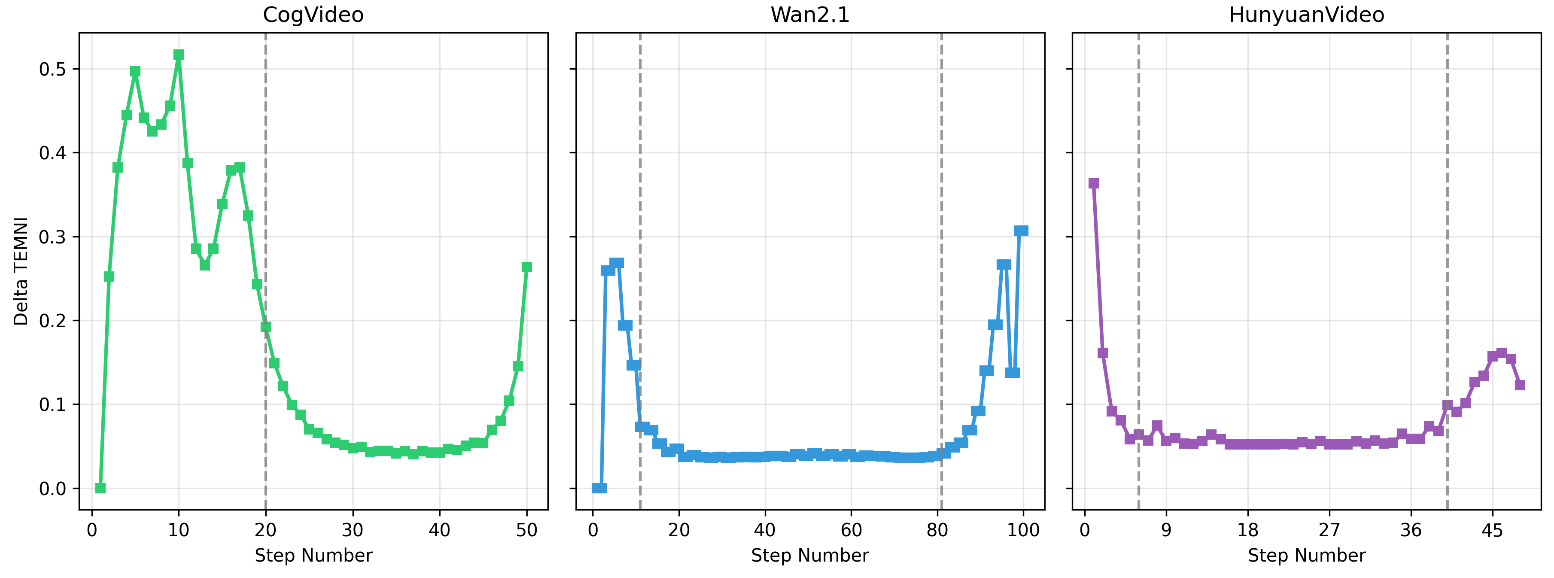}
    \caption{$\Delta$TEMNI signal across all three models throughout the denoising process. Vertical dashed lines mark the transition points identified in the calibration study.}
    \label{fig:metric-fd}
\end{figure}

\begin{figure}[p]
    \centering
    \includegraphics[width=1\linewidth]{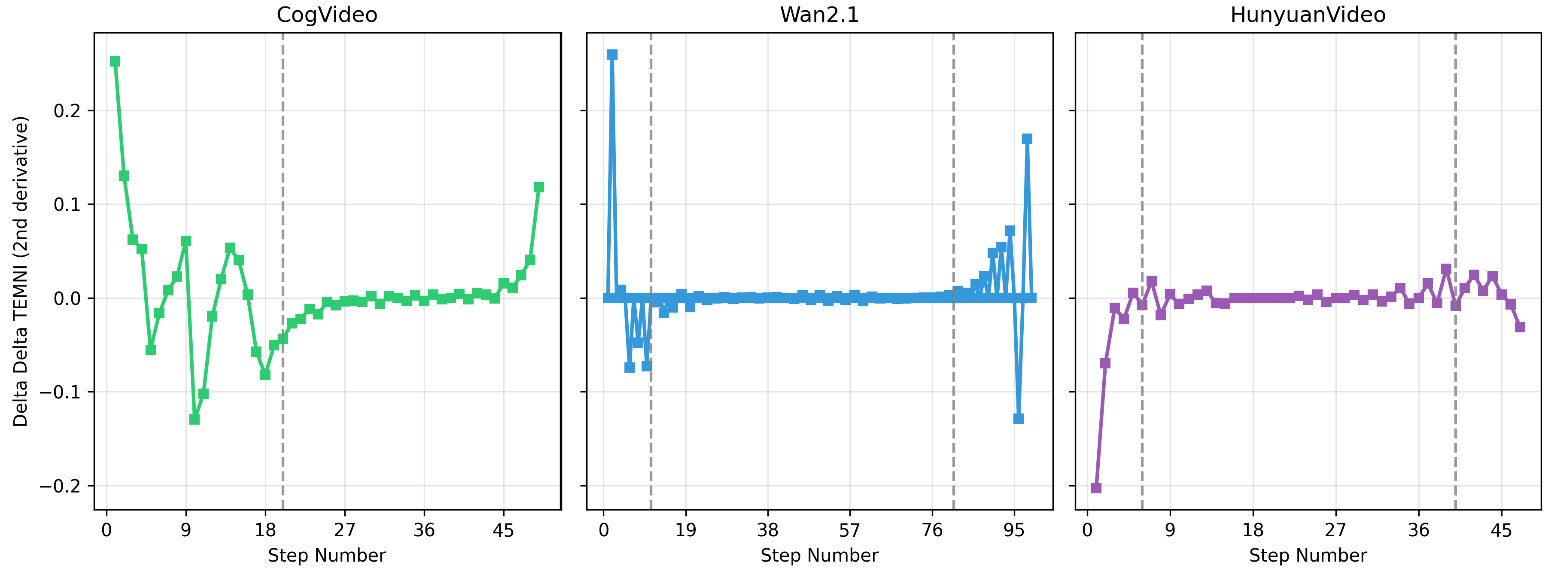}
    \caption{Second derivative of the $\Delta$TEMNI signal across all three models, with the same transition markers. The transition points align with where the second derivative stabilizes, motivating its use as the criterion for placing threshold transitions.}
    \label{fig:metric-sd}
\end{figure}

Figures~\ref{fig:prompt-invariance-appendix}--\ref{fig:metric-sd} support the analysis in Section~\ref{sec:analysis}.

\clearpage

\section{Pareto Frontier Plots}
\label{sec:appendix-pareto}

\begin{figure}[p]
    \centering
    \begin{subfigure}[b]{\linewidth}
        \includegraphics[width=\linewidth]{pareto_plots/pareto_hv_teacache.png}
        \caption{HunyuanVideo}
        \label{fig:pareto-hv-teacache}
    \end{subfigure}
    \begin{subfigure}[b]{\linewidth}
        \includegraphics[width=\linewidth]{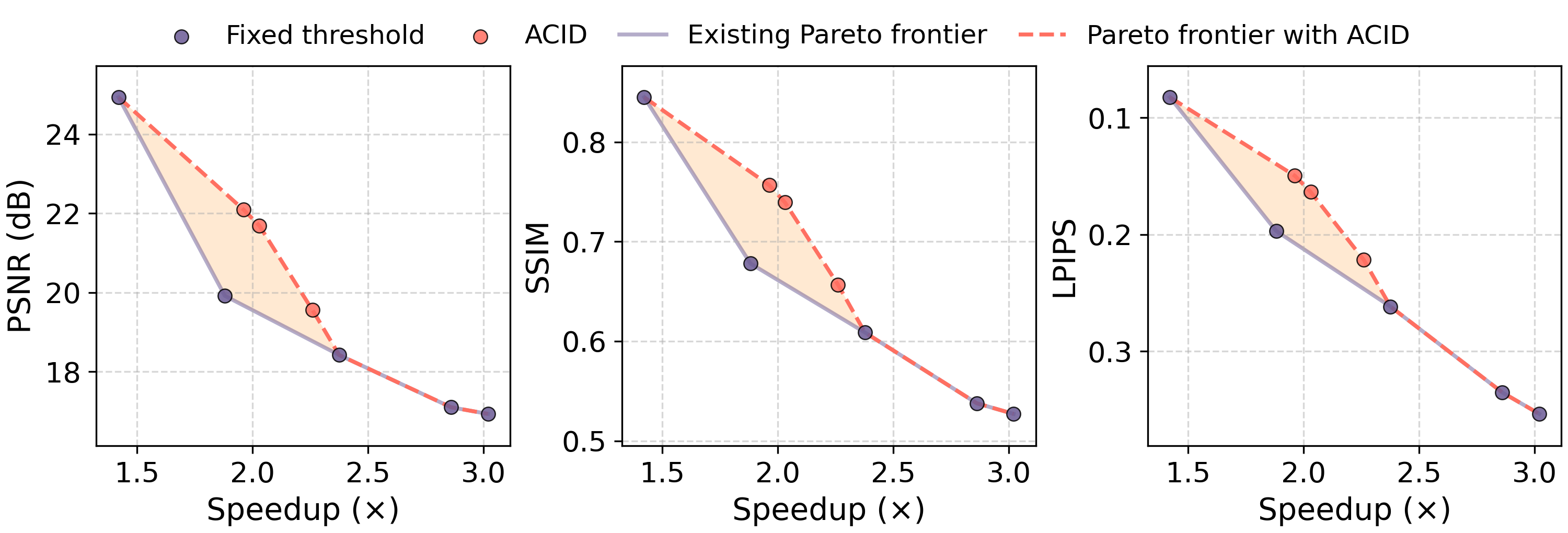}
        \caption{Wan~2.1}
        \label{fig:pareto-wan-teacache}
    \end{subfigure}
    \begin{subfigure}[b]{\linewidth}
        \includegraphics[width=\linewidth]{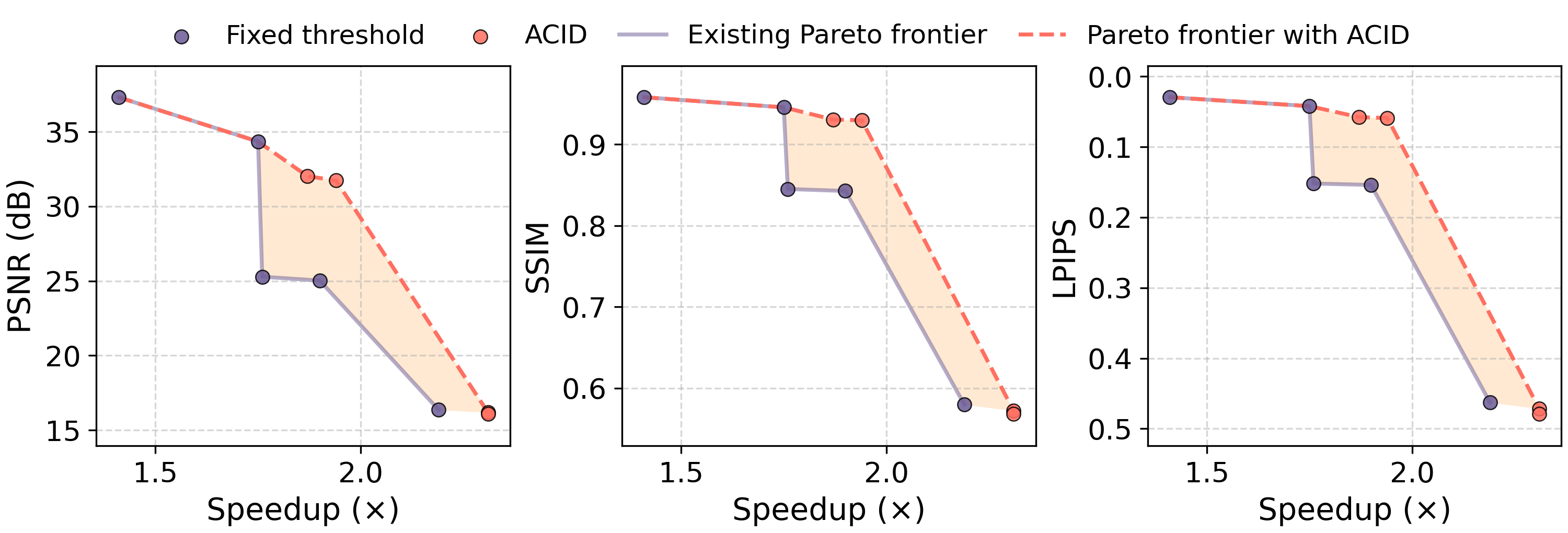}
        \caption{CogVideoX~1.5}
        \label{fig:pareto-cog-teacache}
    \end{subfigure}
    \caption{TeaCache Pareto frontiers across all three models. \system (red markers) expands the frontier beyond any fixed-threshold configuration (purple markers).}
    \label{fig:pareto-teacache}
\end{figure}

\begin{figure}[p]
    \centering
    \begin{subfigure}[b]{\linewidth}
        \includegraphics[width=\linewidth]{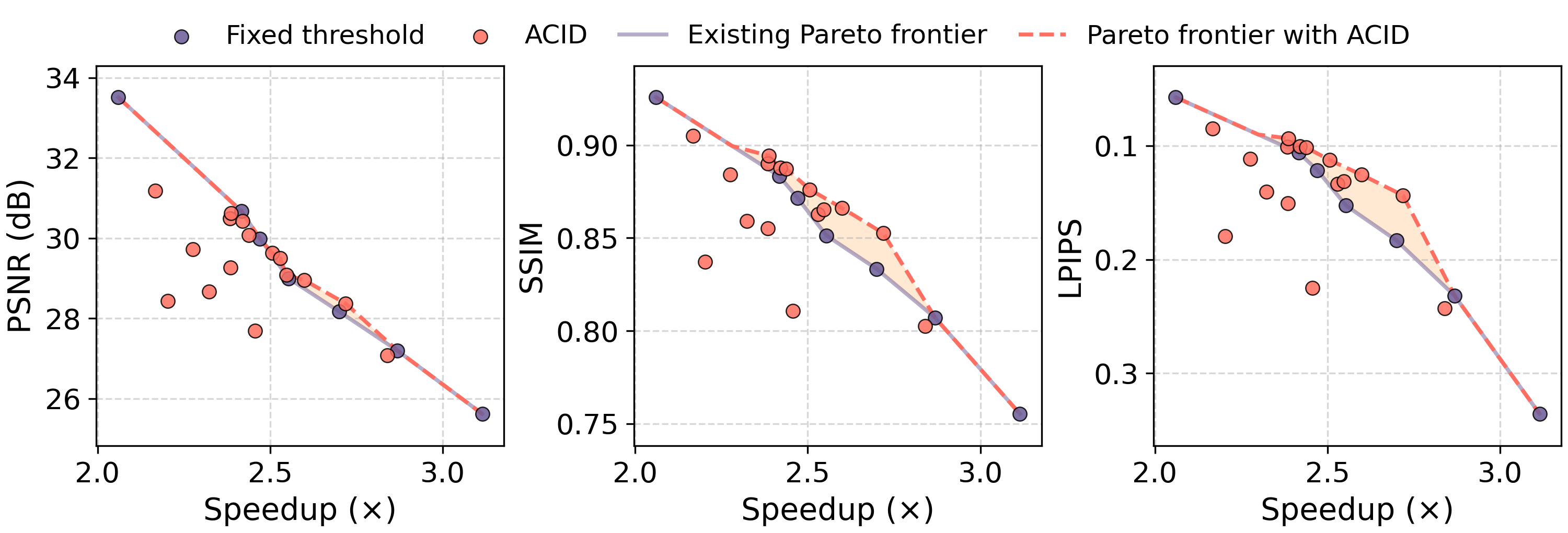}
        \caption{HunyuanVideo}
        \label{fig:pareto-hv-easycache}
    \end{subfigure}
    \begin{subfigure}[b]{\linewidth}
        \includegraphics[width=\linewidth]{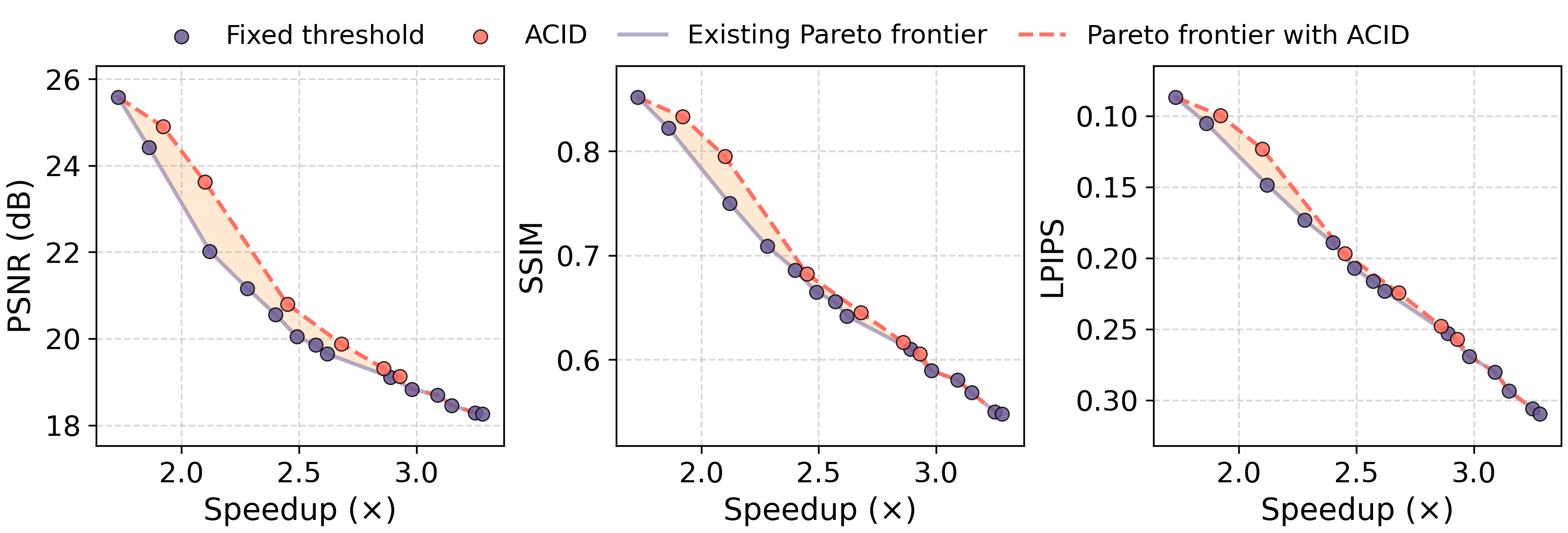}
        \caption{Wan~2.1}
        \label{fig:pareto-wan-easycache}
    \end{subfigure}
    \caption{EasyCache Pareto frontiers. \system expands the frontier beyond any fixed-threshold configuration.}
    \label{fig:pareto-easycache}
\end{figure}

\begin{figure}[p]
    \centering
    \begin{subfigure}[b]{\linewidth}
        \includegraphics[width=\linewidth]{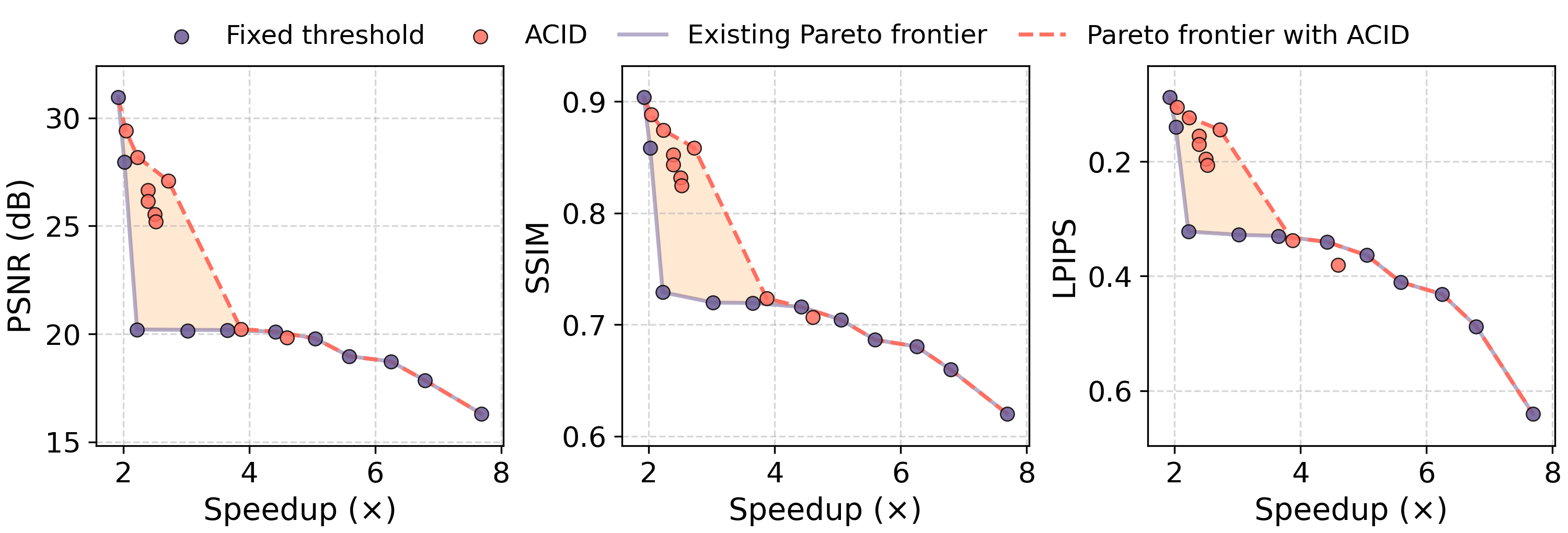}
        \caption{HunyuanVideo}
        \label{fig:pareto-hv-dicache}
    \end{subfigure}
    \begin{subfigure}[b]{\linewidth}
        \includegraphics[width=\linewidth]{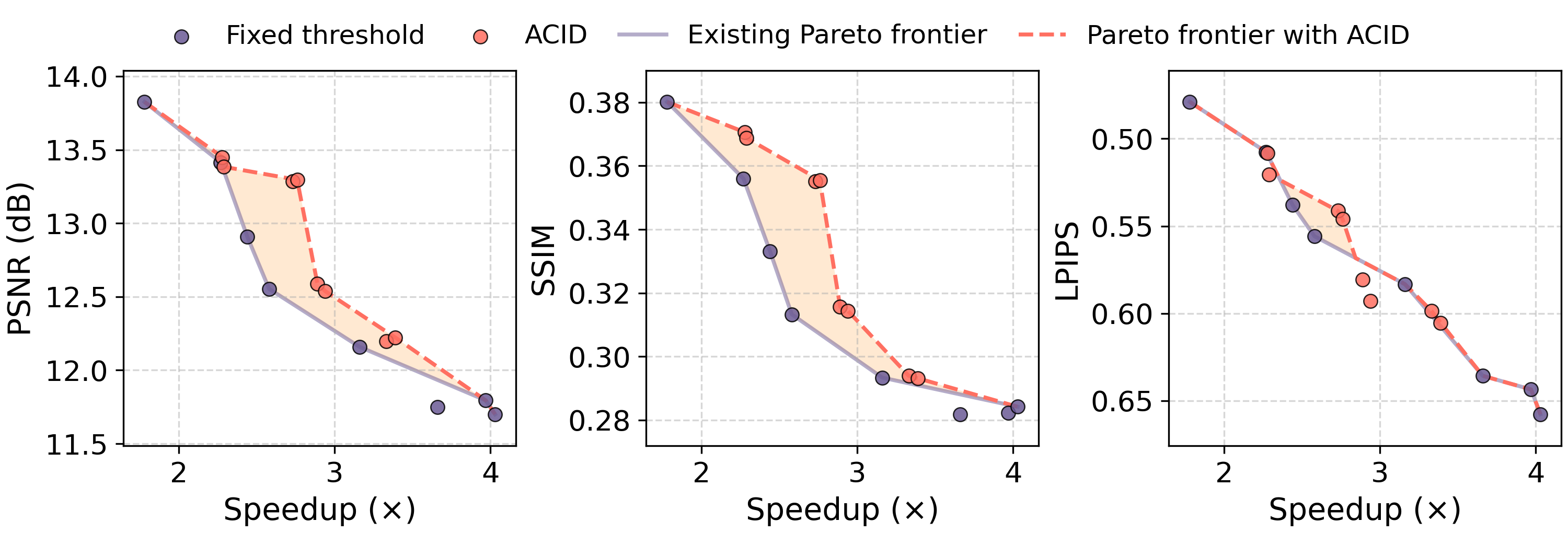}
        \caption{Wan~2.1}
        \label{fig:pareto-wan-dicache}
    \end{subfigure}
    \caption{DiCache Pareto frontiers. \system expands the frontier beyond any fixed-threshold configuration.}
    \label{fig:pareto-dicache}
\end{figure}

Figures~\ref{fig:pareto-teacache}--\ref{fig:pareto-dicache} show Pareto frontier plots for all evaluated method--model combinations. Each plot compares fixed-threshold configurations across a range of thresholds against our scheduled-threshold variants. In each case, scheduled points expand the frontier beyond what any fixed threshold achieves.

\clearpage

\section{Detailed Per-Combo Results}
\label{sec:appendix-detailed}

All configurations evaluated for each method $\times$ model combination are listed
below, sorted by increasing speedup. \textbf{Bold} rows are scheduled modes.
Fidelity metrics (PSNR, SSIM, LPIPS) are computed against the uncached baseline.
$\tau$ denotes the caching threshold for all methods (cache-reuse residual for
TeaCache/EasyCache; feature-difference score for DiCache).
Subscripts $\ell$/$h$ mark the low/high threshold in scheduled modes.
Scheduled rows that share the same threshold pair differ in where the threshold transitions are placed.

\begin{table}[htbp]
\centering
\small
\setlength{\tabcolsep}{5pt}
\caption{TeaCache on HunyuanVideo: all evaluated configurations}
\label{tab:app-tc-hv}
\begin{tabular}{lrrccc}
\toprule
\textbf{Mode} & \textbf{Latency} & \textbf{Speedup} & \textbf{PSNR~$\uparrow$} & \textbf{SSIM~$\uparrow$} & \textbf{LPIPS~$\downarrow$} \\
\midrule
  Fixed ($\tau=0.1$) & 864s & 1.57$\times$ & 24.69 & 0.819 & 0.187 \\
  \textbf{Scheduled ($\tau_\ell=0.1$, $\tau_h=0.25$)} & 654s & 2.08$\times$ & 24.47 & 0.814 & 0.194 \\
  \textbf{Scheduled ($\tau_\ell=0.1$, $\tau_h=0.2$)} & 652s & 2.09$\times$ & 23.85 & 0.796 & 0.215 \\
  \textbf{Scheduled ($\tau_\ell=0.1$, $\tau_h=0.3$)} & 628s & 2.16$\times$ & 24.41 & 0.813 & 0.196 \\
  \textbf{Scheduled ($\tau_\ell=0.15$, $\tau_h=0.3$)} & 554s & 2.45$\times$ & 24.04 & 0.802 & 0.209 \\
  Fixed ($\tau=0.2$) & 542s & 2.51$\times$ & 21.71 & 0.755 & 0.276 \\
  \textbf{Scheduled ($\tau_\ell=0.2$, $\tau_h=0.3$)} & 483s & 2.81$\times$ & 21.66 & 0.754 & 0.281 \\
  Fixed ($\tau=0.3$) & 424s & 3.20$\times$ & 19.98 & 0.693 & 0.353 \\
\bottomrule
\end{tabular}
\end{table}

\begin{table}[htbp]
\centering
\small
\setlength{\tabcolsep}{5pt}
\caption{TeaCache on Wan~2.1: all evaluated configurations}
\label{tab:app-tc-wan}
\begin{tabular}{lrrccc}
\toprule
\textbf{Mode} & \textbf{Latency} & \textbf{Speedup} & \textbf{PSNR~$\uparrow$} & \textbf{SSIM~$\uparrow$} & \textbf{LPIPS~$\downarrow$} \\
\midrule
  Fixed ($\tau=0.05$) & 150s & 1.42$\times$ & 24.93 & 0.845 & 0.083 \\
  Fixed ($\tau=0.1$) & 113s & 1.88$\times$ & 19.92 & 0.678 & 0.197 \\
  \textbf{Scheduled ($\tau_\ell=0.05$, $\tau_h=0.2$)} & 108s & 1.96$\times$ & 22.10 & 0.757 & 0.150 \\
  \textbf{Scheduled ($\tau_\ell=0.05$, $\tau_h=0.25$)} & 105s & 2.03$\times$ & 21.69 & 0.739 & 0.164 \\
  \textbf{Scheduled ($\tau_\ell=0.1$, $\tau_h=0.2$)} & 94s & 2.26$\times$ & 19.56 & 0.657 & 0.222 \\
  Fixed ($\tau=0.15$) & 89s & 2.38$\times$ & 18.42 & 0.609 & 0.262 \\
  Fixed ($\tau=0.2$) & 74s & 2.86$\times$ & 17.10 & 0.538 & 0.335 \\
  Fixed ($\tau=0.25$) & 70s & 3.02$\times$ & 16.94 & 0.527 & 0.354 \\
\bottomrule
\end{tabular}
\end{table}

\begin{table}[htbp]
\centering
\small
\setlength{\tabcolsep}{5pt}
\caption{TeaCache on CogVideoX~1.5: all evaluated configurations}
\label{tab:app-tc-cog}
\begin{tabular}{lrrccc}
\toprule
\textbf{Mode} & \textbf{Latency} & \textbf{Speedup} & \textbf{PSNR~$\uparrow$} & \textbf{SSIM~$\uparrow$} & \textbf{LPIPS~$\downarrow$} \\
\midrule
  Fixed ($\tau=0.1$) & 723s & 1.41$\times$ & 37.32 & 0.958 & 0.029 \\
  Fixed ($\tau=0.2$) & 582s & 1.75$\times$ & 34.35 & 0.945 & 0.042 \\
  Fixed ($\tau=0.22$) & 579s & 1.76$\times$ & 25.28 & 0.845 & 0.152 \\
  \textbf{Scheduled ($\tau_\ell=0.1$, $\tau_h=0.3$)} & 545s & 1.87$\times$ & 32.02 & 0.930 & 0.058 \\
  Fixed ($\tau=0.25$) & 536s & 1.90$\times$ & 25.03 & 0.843 & 0.154 \\
  \textbf{Scheduled ($\tau_\ell=0.1$, $\tau_h=0.3$)} & 525s & 1.94$\times$ & 31.74 & 0.929 & 0.059 \\
  Fixed ($\tau=0.3$) & 465s & 2.19$\times$ & 16.35 & 0.580 & 0.463 \\
  \textbf{Scheduled ($\tau_\ell=0.30$, $\tau_h=0.35$)} & 441s & 2.31$\times$ & 16.18 & 0.572 & 0.472 \\
  \textbf{Scheduled ($\tau_\ell=0.30$, $\tau_h=0.40$)} & 441s & 2.31$\times$ & 16.08 & 0.568 & 0.479 \\
\bottomrule
\end{tabular}
\end{table}

\begin{table}[htbp]
\centering
\footnotesize
\setlength{\tabcolsep}{5pt}
\caption{EasyCache on HunyuanVideo: all evaluated configurations}
\label{tab:app-ec-hv}
\begin{tabular}{lrrccc}
\toprule
\textbf{Mode} & \textbf{Latency} & \textbf{Speedup} & \textbf{PSNR~$\uparrow$} & \textbf{SSIM~$\uparrow$} & \textbf{LPIPS~$\downarrow$} \\
\midrule
  Fixed ($\tau=0.025$) & 659s & 2.06$\times$ & 33.51 & 0.926 & 0.058 \\
  \textbf{Scheduled ($\tau_\ell=0.010$, $\tau_h=0.050$)} & 626s & 2.17$\times$ & 31.18 & 0.905 & 0.085 \\
  \textbf{Scheduled ($\tau_\ell=0.010$, $\tau_h=0.120$)} & 616s & 2.20$\times$ & 28.43 & 0.837 & 0.179 \\
  \textbf{Scheduled ($\tau_\ell=0.010$, $\tau_h=0.075$)} & 596s & 2.28$\times$ & 29.72 & 0.884 & 0.112 \\
  \textbf{Scheduled ($\tau_\ell=0.010$, $\tau_h=0.100$)} & 584s & 2.32$\times$ & 28.67 & 0.859 & 0.141 \\
  \textbf{Scheduled ($\tau_\ell=0.025$, $\tau_h=0.075$)} & 569s & 2.38$\times$ & 30.49 & 0.890 & 0.101 \\
  \textbf{Scheduled ($\tau_\ell=0.025$, $\tau_h=0.090$)} & 569s & 2.38$\times$ & 29.27 & 0.855 & 0.151 \\
  \textbf{Scheduled ($\tau_\ell=0.025$, $\tau_h=0.045$)} & 568s & 2.39$\times$ & 30.62 & 0.894 & 0.094 \\
  Fixed ($\tau=0.0375$) & 561s & 2.42$\times$ & 30.67 & 0.883 & 0.106 \\
  \textbf{Scheduled ($\tau_\ell=0.025$, $\tau_h=0.050$)} & 561s & 2.42$\times$ & 30.43 & 0.888 & 0.101 \\
  \textbf{Scheduled ($\tau_\ell=0.025$, $\tau_h=0.050$)} & 557s & 2.44$\times$ & 30.08 & 0.887 & 0.102 \\
  \textbf{Scheduled ($\tau_\ell=0.025$, $\tau_h=0.150$)} & 552s & 2.46$\times$ & 27.69 & 0.811 & 0.225 \\
  Fixed ($\tau=0.040$) & 549s & 2.47$\times$ & 29.98 & 0.872 & 0.122 \\
  \textbf{Scheduled ($\tau_\ell=0.030$, $\tau_h=0.050$)} & 542s & 2.51$\times$ & 29.63 & 0.876 & 0.113 \\
  \textbf{Scheduled ($\tau_\ell=0.0375$, $\tau_h=0.050$)} & 537s & 2.53$\times$ & 29.50 & 0.863 & 0.134 \\
  \textbf{Scheduled ($\tau_\ell=0.025$, $\tau_h=0.075$)} & 533s & 2.55$\times$ & 29.08 & 0.865 & 0.132 \\
  Fixed ($\tau=0.045$) & 531s & 2.55$\times$ & 28.99 & 0.851 & 0.153 \\
  \textbf{Scheduled ($\tau_\ell=0.030$, $\tau_h=0.060$)} & 522s & 2.60$\times$ & 28.95 & 0.866 & 0.126 \\
  Fixed ($\tau=0.050$) & 503s & 2.70$\times$ & 28.17 & 0.833 & 0.183 \\
  \textbf{Scheduled ($\tau_\ell=0.035$, $\tau_h=0.065$)} & 499s & 2.72$\times$ & 28.37 & 0.853 & 0.144 \\
  \textbf{Scheduled ($\tau_\ell=0.050$, $\tau_h=0.120$)} & 478s & 2.84$\times$ & 27.08 & 0.803 & 0.243 \\
  Fixed ($\tau=0.060$) & 473s & 2.87$\times$ & 27.20 & 0.807 & 0.232 \\
  Fixed ($\tau=0.075$) & 436s & 3.12$\times$ & 25.62 & 0.755 & 0.336 \\
\bottomrule
\end{tabular}
\end{table}

\begin{table}[htbp]
\centering
\footnotesize
\setlength{\tabcolsep}{5pt}
\caption{EasyCache on Wan~2.1: all evaluated configurations}
\label{tab:app-ec-wan}
\begin{tabular}{lrrccc}
\toprule
\textbf{Mode} & \textbf{Latency} & \textbf{Speedup} & \textbf{PSNR~$\uparrow$} & \textbf{SSIM~$\uparrow$} & \textbf{LPIPS~$\downarrow$} \\
\midrule
  Fixed ($\tau=0.020$) & 125s & 1.73$\times$ & 25.58 & 0.852 & 0.087 \\
  Fixed ($\tau=0.025$) & 116s & 1.86$\times$ & 24.42 & 0.822 & 0.105 \\
  \textbf{Scheduled ($\tau_\ell=0.020$, $\tau_h=0.040$)} & 112s & 1.92$\times$ & 24.90 & 0.833 & 0.100 \\
  \textbf{Scheduled ($\tau_\ell=0.025$, $\tau_h=0.050$)} & 103s & 2.10$\times$ & 23.62 & 0.795 & 0.123 \\
  Fixed ($\tau=0.040$) & 102s & 2.12$\times$ & 22.02 & 0.750 & 0.148 \\
  Fixed ($\tau=0.050$) & 95s & 2.28$\times$ & 21.16 & 0.709 & 0.173 \\
  Fixed ($\tau=0.060$) & 90s & 2.40$\times$ & 20.55 & 0.686 & 0.189 \\
  \textbf{Scheduled ($\tau_\ell=0.020$, $\tau_h=0.025$)} & 88s & 2.45$\times$ & 20.80 & 0.682 & 0.197 \\
  Fixed ($\tau=0.070$) & 87s & 2.49$\times$ & 20.04 & 0.665 & 0.207 \\
  Fixed ($\tau=0.080$) & 84s & 2.57$\times$ & 19.85 & 0.656 & 0.216 \\
  Fixed ($\tau=0.090$) & 82s & 2.62$\times$ & 19.64 & 0.642 & 0.223 \\
  \textbf{Scheduled ($\tau_\ell=0.020$, $\tau_h=0.050$)} & 81s & 2.68$\times$ & 19.88 & 0.645 & 0.224 \\
  \textbf{Scheduled ($\tau_\ell=0.020$, $\tau_h=0.075$)} & 76s & 2.86$\times$ & 19.31 & 0.617 & 0.248 \\
  Fixed ($\tau=0.125$) & 75s & 2.89$\times$ & 19.11 & 0.610 & 0.253 \\
  \textbf{Scheduled ($\tau_\ell=0.020$, $\tau_h=0.1$)} & 74s & 2.93$\times$ & 19.13 & 0.605 & 0.257 \\
  Fixed ($\tau=0.150$) & 72s & 2.98$\times$ & 18.83 & 0.589 & 0.269 \\
  Fixed ($\tau=0.175$) & 70s & 3.09$\times$ & 18.70 & 0.581 & 0.280 \\
  Fixed ($\tau=0.200$) & 69s & 3.15$\times$ & 18.45 & 0.568 & 0.293 \\
  Fixed ($\tau=0.225$) & 66s & 3.25$\times$ & 18.29 & 0.550 & 0.306 \\
  Fixed ($\tau=0.250$) & 66s & 3.28$\times$ & 18.26 & 0.548 & 0.310 \\
\bottomrule
\end{tabular}
\end{table}

\begin{table}[htbp]
\centering
\footnotesize
\setlength{\tabcolsep}{5pt}
\caption{DiCache on HunyuanVideo: all evaluated configurations}
\label{tab:app-dc-hv}
\begin{tabular}{lrrccc}
\toprule
\textbf{Mode} & \textbf{Latency} & \textbf{Speedup} & \textbf{PSNR~$\uparrow$} & \textbf{SSIM~$\uparrow$} & \textbf{LPIPS~$\downarrow$} \\
\midrule
  Fixed ($\tau=0.05$) & 659s & 1.92$\times$ & 30.96 & 0.904 & 0.088 \\
  Fixed ($\tau=0.07$) & 626s & 2.02$\times$ & 27.95 & 0.858 & 0.140 \\
  \textbf{Scheduled ($\tau_\ell=0.05$, $\tau_h=0.10$)} & 620s & 2.04$\times$ & 29.42 & 0.888 & 0.105 \\
  Fixed ($\tau=0.08$) & 570s & 2.22$\times$ & 20.21 & 0.729 & 0.322 \\
  \textbf{Scheduled ($\tau_\ell=0.05$, $\tau_h=0.15$)} & 567s & 2.23$\times$ & 28.19 & 0.874 & 0.123 \\
  \textbf{Scheduled ($\tau_\ell=0.05$, $\tau_h=0.25$)} & 529s & 2.39$\times$ & 26.66 & 0.852 & 0.155 \\
  \textbf{Scheduled ($\tau_\ell=0.05$, $\tau_h=0.30$)} & 529s & 2.39$\times$ & 26.15 & 0.844 & 0.169 \\
  \textbf{Scheduled ($\tau_\ell=0.05$, $\tau_h=0.35$)} & 506s & 2.50$\times$ & 25.55 & 0.832 & 0.195 \\
  \textbf{Scheduled ($\tau_\ell=0.05$, $\tau_h=0.40$)} & 502s & 2.52$\times$ & 25.20 & 0.824 & 0.206 \\
  \textbf{Scheduled ($\tau_\ell=0.05$, $\tau_h=0.20$)} & 465s & 2.72$\times$ & 27.08 & 0.859 & 0.144 \\
  Fixed ($\tau=0.10$) & 419s & 3.02$\times$ & 20.15 & 0.720 & 0.328 \\
  Fixed ($\tau=0.15$) & 347s & 3.65$\times$ & 20.18 & 0.719 & 0.330 \\
  \textbf{Scheduled ($\tau_\ell=0.10$, $\tau_h=0.30$)} & 327s & 3.87$\times$ & 20.23 & 0.724 & 0.338 \\
  Fixed ($\tau=0.20$) & 286s & 4.42$\times$ & 20.11 & 0.716 & 0.340 \\
  \textbf{Scheduled ($\tau_\ell=0.15$, $\tau_h=0.40$)} & 275s & 4.60$\times$ & 19.83 & 0.707 & 0.380 \\
  Fixed ($\tau=0.25$) & 250s & 5.05$\times$ & 19.79 & 0.704 & 0.364 \\
  Fixed ($\tau=0.30$) & 226s & 5.59$\times$ & 18.97 & 0.686 & 0.410 \\
  Fixed ($\tau=0.35$) & 202s & 6.25$\times$ & 18.73 & 0.680 & 0.431 \\
  Fixed ($\tau=0.40$) & 186s & 6.79$\times$ & 17.85 & 0.660 & 0.488 \\
  Fixed ($\tau=0.60$) & 164s & 7.69$\times$ & 16.30 & 0.620 & 0.641 \\
\bottomrule
\end{tabular}
\end{table}

\begin{table}[htbp]
\centering
\footnotesize
\setlength{\tabcolsep}{5pt}
\caption{DiCache on Wan~2.1: all evaluated configurations}
\label{tab:app-dc-wan}
\begin{tabular}{lrrccc}
\toprule
\textbf{Mode} & \textbf{Latency} & \textbf{Speedup} & \textbf{PSNR~$\uparrow$} & \textbf{SSIM~$\uparrow$} & \textbf{LPIPS~$\downarrow$} \\
\midrule
  Fixed ($\tau=0.05$) & 126s & 1.78$\times$ & 13.83 & 0.380 & 0.479 \\
  Fixed ($\tau=0.07$) & 99s & 2.27$\times$ & 13.41 & 0.356 & 0.508 \\
  \textbf{Scheduled ($\tau_\ell=0.05$, $\tau_h=0.225$)} & 98s & 2.28$\times$ & 13.45 & 0.371 & 0.508 \\
  \textbf{Scheduled ($\tau_\ell=0.05$, $\tau_h=0.25$)} & 98s & 2.29$\times$ & 13.38 & 0.369 & 0.520 \\
  Fixed ($\tau=0.08$) & 92s & 2.44$\times$ & 12.91 & 0.333 & 0.538 \\
  Fixed ($\tau=0.10$) & 87s & 2.58$\times$ & 12.55 & 0.313 & 0.556 \\
  \textbf{Scheduled ($\tau_\ell=0.07$, $\tau_h=0.225$)} & 82s & 2.73$\times$ & 13.29 & 0.355 & 0.541 \\
  \textbf{Scheduled ($\tau_\ell=0.07$, $\tau_h=0.25$)} & 81s & 2.76$\times$ & 13.30 & 0.356 & 0.546 \\
  \textbf{Scheduled ($\tau_\ell=0.10$, $\tau_h=0.225$)} & 78s & 2.89$\times$ & 12.59 & 0.316 & 0.581 \\
  \textbf{Scheduled ($\tau_\ell=0.10$, $\tau_h=0.25$)} & 76s & 2.94$\times$ & 12.54 & 0.314 & 0.593 \\
  Fixed ($\tau=0.15$) & 71s & 3.16$\times$ & 12.16 & 0.293 & 0.583 \\
  \textbf{Scheduled ($\tau_\ell=0.15$, $\tau_h=0.225$)} & 67s & 3.33$\times$ & 12.20 & 0.294 & 0.599 \\
  \textbf{Scheduled ($\tau_\ell=0.15$, $\tau_h=0.25$)} & 66s & 3.39$\times$ & 12.22 & 0.293 & 0.605 \\
  Fixed ($\tau=0.20$) & 61s & 3.66$\times$ & 11.75 & 0.282 & 0.636 \\
  Fixed ($\tau=0.225$) & 56s & 3.97$\times$ & 11.79 & 0.282 & 0.644 \\
  Fixed ($\tau=0.25$) & 56s & 4.03$\times$ & 11.70 & 0.284 & 0.658 \\
\bottomrule
\end{tabular}
\end{table}

\end{document}